\RequirePackage{fix-cm}
\documentclass[twocolumn]{svjour3}       
\smartqed  

\usepackage{float}
\usepackage{graphicx}
\usepackage{natbib}
\usepackage{graphicx}
\usepackage{latexsym}
\usepackage{algorithm2e}
\usepackage{algpseudocode} 
\usepackage[utf8]{inputenc}
\usepackage[T1]{fontenc}
\usepackage[linguistics]{forest}
\usepackage{tikz}
\usetikzlibrary{calc, shapes, backgrounds}
\usepackage{amsmath, amssymb}
\usepackage[euler]{textgreek}
\usepackage{hyperref}
\usepackage{multirow}
\usepackage{xcolor}

\begin{document}

\title{Bayesian Additive Regression Trees with Model Trees
}


\author{Estev\~ao B. Prado \and
        Rafael A. Moral \and 
        Andrew C. Parnell 
}


\institute{Estev\~ao B. Prado \at
              Hamilton Institute, Insight Centre for Data Analytics, and Department of Mathematics and Statistics, Maynooth University, Ireland \\
              \email{estevao.prado@mu.ie}           
           \and
           Rafael A. Moral \at
              Hamilton Institute and Department of Mathematics and Statistics, Maynooth University, Ireland
            \and
           Andrew C. Parnell \at
              Hamilton Institute, Insight Centre for Data Analytics, and Department of Mathematics and Statistic, Maynooth University, Ireland
}

\date{Updated: January 9th, 2021.}

\maketitle

\begin{abstract}
Bayesian Additive Regression Trees (BART) is a tree-based machine learning method that has been successfully applied to regression and classification problems. BART assumes regularisation priors on a set of trees that work as weak learners and is very flexible for predicting in the presence of non-linearity and high-order interactions. In this paper, we introduce an extension of BART, called Model Trees BART (MOTR-BART), that considers piecewise linear functions at node levels instead of piecewise constants. In MOTR-BART, rather than having a unique value at node level for the prediction, a linear predictor is estimated considering the covariates that have been used as the split variables in the corresponding tree. In our approach, local linearities are captured more efficiently and fewer trees are required to achieve equal or better performance than BART. Via simulation studies and real data applications, we compare MOTR-BART to its main competitors. R code for MOTR-BART implementation is available at \url{https://github.com/ebprado/MOTR-BART}.

\keywords{Bayesian Trees \and Linear models \and Machine Learning \and Bayesian Non-parametric regression}
\end{abstract}

\section{Introduction}
\label{intro}

Bayesian Additive Regression Trees (BART) is a statistical method proposed by \cite{chipman2010bart} that has become popular in recent years due to its competitive performance on regression and classification problems, when compared to other supervised machine learning methods, such as Random Forests (RF) \citep{breiman2001random} and Gradient Boosting (GB) \citep{friedman2001greedy}. BART differs from other tree-based methods as it controls the structure of each tree via a prior distribution and generates the predictions via an MCMC backfitting algorithm that is responsible for accepting and rejecting the proposed trees along the iterations. In practice, BART can be used for predicting a continuous/binary response variable through R packages, such as \texttt{dbarts} \citep{chipman2010bart}, \texttt{BART} \citep{BARTpackage} and \texttt{bartMachine} \citep{bartMachine}.

In essence, BART is a non-parametric Bayesian algorithm that generates a set of trees by choosing the covariates and the split-points at random. To generate the predicted values for each terminal node, the Normal distribution is adopted as the likelihood function as well as prior distributions are placed on the trees, predicted values and variance of the predictions. Through a backfitting MCMC algorithm, the predictions from each tree are obtained by combining Gibbs Sampler and Metropolis-Hastings steps. The final prediction is then calculated as the sum of the predicted values over all trees. In parallel,
samples from the posterior distributions of the quantities of interest are naturally generated along the MCMC iterations.

In this paper, we introduce the algorithm MOTR-BART, which combines Model Trees \citep{quinlan1992learning} with BART to deal with local linearity at node levels. In MOTR-BART, rather than estimating a constant as the predicted value as BART does, for each terminal node a linear predictor is estimated, including only the covariates that have been used as a split in the corresponding tree. With this approach, we aim to capture linear associations between the response and covariates and then improve the final prediction. We observe that MOTR-BART requires fewer trees to achieve equal or better performance than BART as well as reaches faster convergence, when we look at the overall log-likelihood. Through simulation experiments that consider different number of observations and covariates, MOTR-BART outperforms its main competitors in terms of RMSE on out-of-sample data, even using fewer trees. In the real data applications, MOTR-BART is competitive compared to BART and other tree-based methods.

This paper is organised as follows. In Section 2, we briefly introduce BART, some related works, and Model Trees. Section 3 presents the mathematical details of BART and how it may be implemented in the regression context. In Section 4, we introduce the MOTR-BART, providing the mathematical expressions needed for regression and classification. Section 5 shows comparisons between MOTR-BART and other algorithms via simulated scenarios and real data applications. Finally, in Section 6, we conclude with a discussion.


\section{Tree-based methods}
\label{Tree_methods}

\subsection{Related works}
BART considers that a univariate response variable can be approximated by a sum of predicted values from a set of trees as $\hat{y} = \sum_{t=1}^{m} g(\textbf{X}; M_{t}, T_{t})$, where $g(\cdot)$ is a function that assigns a predicted value based on $\textbf{X}$ and $T_{t}$, $\textbf{X}$ is the design matrix, $M_{t}$ the set of predicted values of the tree $t$, and $T_{t}$ represents the structure of the tree $t$. In BART, a tree $T_{t}$ can be modified using four moves (growing, pruning, changing, or swapping), and the splitting rules that create the terminal/internal nodes are randomly chosen. To sample from the full conditional distribution of $T_{t}$, the Metropolis-Hastings algorithm is used. Further, each component $\mu_{t \ell} \in M_{t}$ is sampled from its full conditional via a Gibbs Sampler step. Then, the final prediction is calculated by adding up the values of $\mu_{t \ell}$ from all the $m$ trees. Further details are given in Section \ref{bart_maths}.

BART’s versatility has made it an attractive option with applications in credit risk modelling \citep{zhang2010bayesian}, identification of subgroup effects in clinical trials \citep{sivaganesan2017subgroup, schnell2016bayesian}, competing risk analysis \citep{sparapani2019nonparametric}, survival analysis of stem cell transplantation \citep{sparapani2016nonparametric}, proteomic biomarker discovery \citep{hernandez2015bayesian} and causal inference \citep{hill2011bayesian, green2012modeling, hahn2020bayesian}. In this context, many extensions have been proposed, such as BART for estimating monotone and smooth surfaces \citep{starling2019monotone, starling2020bart, lineroAnDyang2018}, categorical and multinomial data \citep{jared2017, kindo2016multinomial}, high-dimensional data \citep{hernandez2018bayesian, he2018xbart, linero2018bayesian}, zero-inflated and semi-continuous responses \citep{linero2018semiparametric}, heterocedastic data \citep{pratola2017heteroscedastic}, BART with quantile regression and varying coefficient models \citep{kindo2016bayesian,deshpande2020vc}, among others. Recently, some papers have developed theoretical aspects related to BART \citep{linero2017review, rockova2017posterior, rockova2018theory, lineroAnDyang2018}.

Some of the works mentioned above are somewhat related to MOTR-BART. For instance, \cite{lineroAnDyang2018} introduce the soft BART in order to provide an approach suitable for both estimating a target smooth function and dealing with sparsity. In soft BART, the observations are not allocated deterministically to the terminal nodes, as it is commonly done in the conventional trees. Instead, the observations are assigned to the terminal nodes based on a probability measure, which is a function of a bandwidth parameter and of the distance between the values of the covariates and the cut-offs defined by the splitting rules. Through empirical and theoretical results, they show that soft BART is capable to smoothly approximate linear and non-linear functions as well as that its posterior distribution concentrates, under mild conditions, at the minimax rate. The main differences between MOTR-BART and soft BART are: i) MOTR-BART does not use the idea of soft trees, where the observations are assigned to the terminal based on a probability measure, and ii) MOTR-BART uses a linear predictor rather than a piecewise constant to generate the predictions at node-level.

In this sense, \cite{starling2020bart} propose the BART with Targeted Smoothing (tsBART) by introducing smoothness over a covariate of interest. In their approach, rather than predicting a piecewise constant as the standard BART, univariate smooth functions of a certain covariate of interest are used to generate the node-level predictions. In tsBART, they place a Gaussian process prior over the smooth function associated to each terminal node and grow the trees using all available covariates, apart from the one over which they wish to introduce the smoothness. Although tsBART and MOTR-BART have some similarities, since both do not base their predictions on piecewise constants and both aim to provide more flexibility at the node-level predictions, they differ as MOTR-BART allows for more than one covariate to be used in the linear predictors and does not assume a Gaussian process prior on each linear predictor.

In addition, \cite{deshpande2020vc} propose an extension named VCBART that combines varying coefficient models and BART. In their approach, rather than approximating the response variable itself, each covariate effect in the linear predictor is estimated by using BART. They also provide theoretical results about the near minimax optimal rate associated to the posterior concentration of the VCBART considering non-i.i.d errors. Although the linear model is a particular case of the varying coefficients model, MOTR-BART and VCBART are structurally different. For instance, VCBART considers that a univariate response variable can be approximated via an overall linear predictor in which the coefficients are estimated via BART. In contrast, MOTR-BART approximates the response by estimating a linear predictor for each terminal node in each tree, where Normal priors are placed on the coefficients in order to estimate them.

Regarding non-Bayesian methods, we highlight the algorithms introduced by \cite{friedberg2018local} and \cite{kunzel2019linear}, named Local Linear Forests (LLF) and Linear Random Forests (LRF), respectively. In their work, RF-based algorithms are proposed, where the predictions for each terminal node are generated from a local ridge regression. Furthermore, the LLF algorithm also provides a point-wise confidence interval based on the RF delta method proposed by \cite{athey2019generalized} and theoretical results related to asymptotic consistency and rates of convergence of the forest.

\subsection{Model trees}

\cite{quinlan1992learning} introduced the term Model Trees when proposing the M5 algorithm, which is a tree-based method that estimates a linear equation for each terminal node and then computes the final prediction based on piecewise linear models and a smoothing process. Initially introduced in the context of regression, extensions and generalisations for classification were presented by  \cite{wang1997inducing} and \cite{landwehr2005logistic}.

Unlike BART, RF and GB, where multiple trees are generated to predict the outcome, the algorithm M5 generates only one tree. For the growing process, the Variance Reduction (VR) is adopted as the splitting criterion. When estimating the coefficients for the linear equation at a terminal node, the covariates are selected based on tests and, depending on their significance, the linear equation can be reduced to a constant, if all covariates do not show any significance. At the end, the prediction is calculated based on the linear predictors from all terminal nodes and then is averaged over the predictions from the terminal nodes along the path to the root.


\section{BART}\label{bart_maths}

Introduced by \cite{chipman2010bart}, BART is a tree-based machine learning method that considers that a univariate response variable $\textbf{y} = (y_{1},....,y_{n})^{\top}$ can be approximated by a sum-of-trees as
$$
\begin{aligned}
y_{i} = \sum_{t = 1}^{m} g(\textbf{x}_{i}; T_{t}, M_{t}) + \epsilon_{i}, \mbox{ } \epsilon_{i} \sim \mbox{N}(0, \sigma^{2}),
\end{aligned}
$$
where $g(\textbf{x}_{i}; T_{t}, M_{t})$ is a function that assigns a predicted value $\mu_{t \ell}$ based on $\textbf{x}_{i}$, $\textbf{x}_{i} = (x_{i1}, ..., x_{id})$ represents the $i$-th row of the design matrix $\textbf{X}$, $T_{t}$ is the set of splitting rules that defines the $t$-th tree and $M_{t} = (\mu_{t1}, ..., \mu_{t b_{t}})$ is the set of predicted values for all nodes in the tree $t$, with $\mu_{t b_{t}}$ representing the predicted value for the terminal node $b_{t}$. The splitting rules that define the terminal nodes for the tree $t$ can be defined as partitions $\mathcal{P}_{t \ell}$, with $\ell = 1, ..., b_{t}$, and $g(\textbf{x}_{i}; T_{t}, M_{t}) = \mu_{t \ell}$ for all observations $i \in \mathcal{P}_{t \ell}$, based on the values of $\textbf{x}_{i}$.

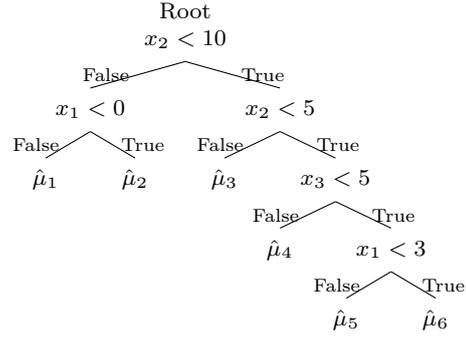
\begin{figure}
\centering
\begin{forest}
for tree={s sep=6mm}
[Root \\ $x_{2} < 10$
[$x_{1} < 0$, edge label={node[midway,left,font=\scriptsize]{False}} [$\hat{\mu}_{1}$, edge label={node[midway,left,font=\scriptsize]{False}}] [$\hat{\mu}_{2}$ , edge label={node[midway,right,font=\scriptsize]{True}}]] [$x_{2} < 5$, edge label={node[midway,right,font=\scriptsize]{True}}
      [$\hat{\mu}_{3}$, edge label={node[midway,left,font=\scriptsize]{False}}]
      [$x_{3} < 5$, edge label={node[midway,right,font=\scriptsize]{True}},
[$\hat{\mu}_{4}$, edge label={node[midway,left,font=\scriptsize]{False}}]
         [$x_{1} < 3$ , edge label={node[midway,right,font=\scriptsize]{True}}, [$\hat{\mu}_{5}$, edge label={node[midway,left,font=\scriptsize]{False}} ] [$\hat{\mu}_{6}$, , edge label={node[midway,right,font=\scriptsize]{True}}]]
      ]
] ]
\end{forest}
\caption{An example of a single tree generated by BART. In practice, BART generates multiple trees for which the predictions are added together. The covariates and split-points that define the terminal nodes are proposed uniformly, and optimised via an MCMC algorithm. The quantities $x_{1}, x_{2}$ and $x_{3}$ represent covariates; $\hat{\mu}_{\ell}$ is the predicted value of node $\ell$.}
\label{BCART_tree}
\end{figure}

In BART, each regression tree is generated as in \cite{chipman1998bayesian} (see Figure \ref{BCART_tree}) where, through a backfitting algorithm, a binary tree can be created or modified by four movements: grow, prune, change or swap. A new tree is created by one of these four movements, and compared to the previous version via a Metropolis-Hastings step on the partial residuals. In the growing process, a terminal node is randomly selected and is separated into two new nodes. Here, the covariate that is used to create the new terminal nodes is picked uniformly as is its associated split-point. In other words, the splitting rule is fully defined assuming the uniform distribution over both the set of covariates and the set of their split-points. During a prune step, a parent of two terminal nodes is randomly chosen and then its child nodes are removed. In the change movement, a pair of terminal nodes is picked at random and its splitting rule is changed. In the swap process, two parents of terminal nodes are randomly selected and their splitting rules are exchanged. 

In order to control the depth of the tree, a regularisation prior is considered as
\begin{align}
\label{Prior_T_t}
p(T_{t}) = \prod_{\ell \in L_{I}} \left[ \alpha (1 + d_{t \ell})^{-\beta} \right] \times \prod_{\ell \in L_{T}} \left[ 1 - \alpha (1 + d_{t \ell})^{-\beta} \right],
\end{align}
where $L_{I}$ and $L_{T}$ denote the sets of indices of the internal and terminal nodes, respectively, $d_{t \ell}$ is the depth of node $\ell$ in tree $t$, $\alpha \in (0,1)$, and $\beta \geq 0$. \cite{chipman2010bart} recommend $\alpha = 0.95$ and $\beta = 2$. In essence, $\alpha (1 + d_{t \ell})^{-\beta}$ computes the probability of the node $\ell$ being internal at depth $d_{t \ell}$.

To estimate the terminal node parameters, $\mu_{t \ell}$, and overall variance, $\sigma^{2}$, conjugate priors are used:
$$
\begin{aligned}
\mu_{t \ell}| T_{t} & \sim \mbox{N}(0, \sigma^{2}_{\mu}), \\
\sigma^{2} & \sim \mbox{IG}(\nu/2, \nu\lambda/2),
\end{aligned}
$$
where $\sigma_{\mu} = 0.5/(c\sqrt{m})$, $1 \leq c \leq 3$, IG denotes the Inverse Gamma distribution, and $m$ is the number of trees. The division by $m$ has the effect of reducing the predictive power of each tree and forcing each to be a weak learner. The joint posterior distribution of the trees and predicted values is given by
$$
\begin{aligned}
p((T,M), \sigma^{2}| \textbf{y}, \textbf{X}) & \propto p(\textbf{y}| \textbf{X}, T, M, \sigma^{2}) p(M|T) p(T) p(\sigma^{2}), \\
& \propto \left[\prod_{t=1}^{m} \prod_{\ell = 1}^{b_{t}} \prod_{i: \textbf{x}_{i} \in \mathcal{P}_{t\ell}} p(y_{i}| \textbf{x}_{i}, T_{t}, M_{t}, \sigma^{2}) \right] \times \\
& \mbox{ } \mbox{ } \mbox{ } \mbox{ }  \times \left[\prod_{t=1}^{m} \prod_{\ell = 1}^{b_{t}} p(\mu_{t \ell}|T_{t}) p(T_{t}) \right] p(\sigma^{2}).
\end{aligned}
$$
\cite{chipman2010bart} initially decompose this joint posterior into two full conditionals. The first one generates all $\mu_{t \ell}$ for each tree $t = 1, ..., m$, and is given by
\begin{align}
\label{BART_full_conditional_T_M}
p(T_{t}, M_{t} &| T_{(-t)}, M_{(-t)}, \sigma^{2}, \textbf{X}, \textbf{y}),
\end{align}
where $T_{(-t)}$ represents the set of all trees without the component $t$; similarly for $M_{(-t)}$. To sample from \eqref{BART_full_conditional_T_M}, \cite{chipman2010bart} noticed that the dependence of the full conditional of $(T_{t}, M_{t})$ on $T_{(-t)}, M_{(-t)}$ is given by the partial residuals through
$$
\begin{aligned}
R_{t} = \textbf{y} - \sum_{k \neq t}^{m} g(\textbf{X}; T_{k}, M_{k}).
\end{aligned}
$$

\noindent
Thus, rather than depending on the other trees and their predicted values, the joint full conditional of $(T_{t}, M_{t})$ may be rewritten as $p(T_{t}, M_{t} | R_{t}, \sigma^{2}, \textbf{X})$, with $R_{t}$ acting like the response variable. This simplification allows us to sample from $p(T_{t}, M_{t} | R_{t}, \sigma^{2}, \textbf{X})$ in two steps:

\begin{itemize}

\item[a)] Propose a new tree either growing, pruning, changing, or swapping terminal nodes via 
$$
\begin{aligned}
p(T_{t}| R_{t}, \sigma^{2}) &  \propto p(T_{t}) \int p(R_{t}| M_{t}, T_{t}, \sigma^{2}) p(M_{t} | T_{t}) dM_{t}, \\
& \propto p(T_{t}) p(R_{t}| T_{t}, \sigma^{2}), \\
& \propto p(T_{t}) \prod_{\ell = 1}^{b_{t}} \left[ \left(\frac{\sigma^{2}}{\sigma^{2}_{\mu} n_{t \ell} + \sigma^{2}}\right)^{1/2} \right. \\
& \left. \mbox{ } \mbox{ } \mbox{ } \mbox{ } \mbox{ } \mbox{ } \mbox{ } \mbox{ } \times \exp \left( \frac{\sigma^{2}_{\mu} \left[ n_{t \ell} \bar{R}_{\ell} \right]^2}{2 \sigma^{2} (\sigma^{2}_{\mu} n_{t \ell} + \sigma^{2})} \right) \right],
\end{aligned}
$$
where $\bar{R}_{\ell} = \sum_{i \in \mathcal{P}_{t \ell}} r_{i} / n_{t \ell}$, $r_{i} \in R_{t}$ and $n_{t \ell}$ is the number of observations that belong to $\mathcal{P}_{t \ell}$. This sampling is carried out through a Metropolis-Hastings step, as the expression does not have a known distributional form;

\vspace{0.5cm}

\item[b)] Generate the predicted values $\mu_{t \ell}$ for all terminal nodes in the corresponding tree. As all $\mu_{t \ell}$ are independent from each other, it is possible to write $p(M_{t}| T_{t}, R_{t}, \sigma^{2}) = \prod_{\ell = 1}^{b_{t}} p(\mu_{t \ell}|T_{t}, R_{t}, \sigma^{2})$. Hence,
$$
\begin{aligned}
p(\mu_{t \ell}| T_{t}, R_{t}, \sigma^{2}) & \propto p(R_{t}| M_{t}, T_{t}, \sigma^{2}) p(\mu_{t \ell}), \\
& \propto \exp \left( - \frac{1}{2 \sigma^{2}_{*}} \left( \mu_{t \ell} - \mu_{t \ell}^{*} \right)^{2}  \right),
\end{aligned}
$$
which is a 
$$
\begin{aligned}
\mbox{N}\left(\frac{ \sigma^{-2}\sum_{i \in \mathcal{P}_{t \ell}} r_{i}}{n_{t \ell}/\sigma^{2} + \sigma^{-2}_{\mu}}, \frac{1}{n_{t \ell}/\sigma^{2} + \sigma^{-2}_{\mu}} \right).
\end{aligned}
$$
\end{itemize}

\noindent
Then, after generating all predicted values for all trees, $\sigma^{2}$ can be updated based on
\begin{align}
\label{BART_full_conditional_sigma2}
p(\sigma^{2} | T, M, \textbf{X}, \textbf{y}) \propto & \mbox{ } p(\textbf{y} | \textbf{X}, T, M, \sigma^{2}) p(\sigma^{2}) \nonumber\\
\propto & \mbox{ } (\sigma^{2})^{-\left(\frac{n + \nu}{2} + 1\right)} \exp \left( -\frac{S + \nu \lambda}{2 \sigma^{2}} \right),
\end{align}
where $S = \sum_{i=1}^{n} (y_{i} - \hat{y}_{i})^{2}$ and $\hat{y}_{i} = \sum_{t = 1}^{m} g(\textbf{x}_{i}; T_{t}, M_{t})$. The expression in \eqref{BART_full_conditional_sigma2} is an $\mbox{IG}((n + \nu)/2, (S + \nu \lambda)/2)$, and drawing samples from it is straightforward.

In Algorithm 1, we present the full structure of the BART algorithm. Firstly, the response variable and design matrix are required. The trees, hyper-parameters, partial residuals and the number of MCMC iterations have to be initialised. Later, within each MCMC iteration, candidate trees ($T^{*}_{t}$) are sequentially generated, which might be accepted (or rejected) as the current trees with probability $\alpha(T_{t}, T^{*}_{t})$. After that, the predicted values $\mu_{t \ell}$ are generated for all terminal nodes and then the partial residuals are updated. Finally, the final predictions and $\sigma^{2}$ are obtained.

\RestyleAlgo{boxruled}
\begin{algorithm}
\label{original_BART_algorithm}
\caption{BART Algorithm}
\SetAlgoLined
\KwResult{A posterior distribution of trees $T$}
\KwData{$\textbf{y}$ (response variable) and $\textbf{X}$ (design matrix);}
 \textbf{Initialise:}
 $T = (T_{1}, ..., T_{m})$ to stumps, $\{\mu_{t \ell}\} = 0$, $R_{1}^{(1)} = \textbf{y}$, $\alpha$, $\beta$, $\sigma^{2}_{\mu}$, $\nu$, $\lambda$, $\sigma^{2}$, number of trees $(m)$ and the number of MCMC iterations (burn-in and post-burn-in) ($nIter$).\\
    
\For{k in 1:nIter}{
    \For{t in 1:m}{
         Propose a new tree $T_{t}^{*}$ by growing, pruning, changing or swapping, where each movement has probability of 0.25 to be chosen;
        
        Compute $\alpha(T_{t}, T_{t}^{*}) = 
        \mbox{min}\left \lbrace 1, 
        \frac{p(R_{t}^{(k)}|T_{t}^{*}, \sigma^{2}) p(T_{t}^{*})}{p(R_{t}^{(k)}|T_{t}, \sigma^{2}) p(T_{t})} \right 
        \rbrace$; 

        Sample $u \sim U[0,1]$; \\
        \textbf{if} $u < \alpha(T_{t}, T_{t}^{*})$ \textbf{then} $T_{t} = T_{t}^{*}$ \textbf{else} $T_{t} = T_{t}$; \\

        \For{$\ell$ in 1:$b_{t}$}{
            Update $\mu_{t \ell}$ from $p(\mu_{t \ell}| T_{t}, R_{t}, \sigma^{2})$;
        }
        Update $R_{t}^{(k)} = \textbf{y} - \sum_{j \neq t}^{m} g(\textbf{X}; T_{j}, M_{j})$;
    }

Update $\hat{\textbf{y}}^{(k)} = \sum_{t = 1}^{m} g(\textbf{X}, T_{t}, M_{t})$;

Update $\sigma^{2}$ sampling from $p(\sigma^{2}|T, M, \textbf{X}, \textbf{y})$.
}
\end{algorithm}


\section{Model Trees BART}
\label{LM_BART}

In MOTR-BART, we consider that the response variable is a sum of trees in the form of
$$\textbf{y} = \sum_{t = 1}^{m} g(\textbf{X}; T_{t}, B_{t}) + \epsilon,$$
where $B_{t}$ is the set of parameters of all linear predictors of the tree $t$. In terms of partial residuals, MOTR-BART can be represented as 
$$
\begin{aligned}
r_{i}|\textbf{x}_{i}, \boldsymbol\beta_{t \ell}, \sigma^{2} & \sim \mbox{N}(\textbf{x}_{i} \boldsymbol\beta_{t \ell}, \sigma^{2}), \\
\end{aligned}
$$
\noindent
where $r_{i} = y_{i} - \sum_{j \neq t}^{m} g(\textbf{x}_{i}; T_{j}, B_{j})$, $\boldsymbol\beta_{t \ell}$ is the parameter vector associated to the terminal node $\ell$ of the tree $t$. In this sense, all observations $i \in \mathcal{P}_{t \ell}$ will have predicted values based on $\boldsymbol\beta_{t \ell}$ and the values of their covariates $\textbf{X}_{t \ell}$. Thus, each observation $i \in \mathcal{P}_{t \ell}$ may have different predicted value.
\noindent
The priors for $\boldsymbol\beta_{t \ell}$ and $\sigma^{2}$ are
\begin{align}
\label{Prior_beta_tl}
\boldsymbol\beta_{t \ell}|T_{t}  & \sim \textbf{N}_{q} (\textbf{0}, \sigma^{2}\textbf{V}),\\
\sigma^{2}_{}|T_{t}   & \sim \mbox{IG}(\nu /2, \nu \lambda /2), \nonumber
\end{align}

\noindent
where $\textbf{V} = \tau_{b}^{-1} \times \textbf{I}_{q}$ and $q = p_{t \ell} + 1$, with $p_{t \ell}$ representing the number of covariates in the linear predictor of the terminal node $\ell$ of the tree $t$. The additional dimension in $\textbf{V}$ is due to a column filled with 1's in the design matrix $\textbf{X}_{t \ell}$. Here, the role of the parameter $\tau_{b}$ is to balance the importance of each tree on the final prediction by keeping the components of $\boldsymbol\beta_{t \ell}$ close to zero, thus avoiding that one tree contributes more than other. Since the prior on the vector $\boldsymbol\beta_{t \ell}$ assumes that all entries have the same variance, we scale the predictors in $\textbf{X}_{t \ell}$ in order to make this assumption valid. In our simulations and real data applications, we have found that $\tau_{b} = m$ worked well.

Another possibility is to penalise the intercept and the slopes differently. In this sense, the specification of intercept- and slope-specific variances may be done by setting $\textbf{V}$ as a $q \times q$ diagonal matrix with $\textbf{V}_{1,1} = \tau^{-1}_{\beta_{0}}$ and $\textbf{V}_{j+1,j+1} = \tau^{-1}_{\beta}$. In addition, we may assume conjugate priors such as $\tau_{\beta_{0}} \sim \mbox{G} (a_{0}, b_{0})$ and $\tau_{\beta} \sim \mbox{G} (a_{1}, b_{1})$ to be able to estimate both variances via Gibbs-sampling steps. In this case, we would end up with the following full conditionals:
\begin{align}
    \tau_{\beta_{0}} | - \sim \mbox{G} \left( \frac{\sum_{t=1}^{m} b_{t}}{2} + a_{0}, \frac{\boldsymbol\beta_{0}^\top \boldsymbol\beta_{0}}{2\sigma^{2}} + b_{0} \right),
    \nonumber
\end{align}
\begin{align}
    \tau_{\beta} | - \sim \mbox{G} \left( \frac{\sum_{t=1}^{m} \sum_{\ell = 1}^{b_{t}} p_{t \ell}}{2} + a_{1}, \frac{\boldsymbol\beta^\top_{*} \boldsymbol\beta_{*}}{2\sigma^{2}} + b_{1} \right),
    \nonumber
\end{align}
where $\boldsymbol\beta_{0}$ is a vector with the intercepts from all terminal nodes of all trees and $\boldsymbol\beta_{*}$ contains the slopes from all linear predictors of all trees. In our software, we have implemented an option, through the argument \texttt{vars\_inter\_slope = TRUE/FALSE}, that allows the user to either estimate $\tau_{\beta_{0}}$ and $\tau_{\beta}$ or use $\tau_{b} = m$. In Section \ref{Results}, we show the results of MOTR-BART using both approaches.

Hence, the full conditionals are
$$
\begin{aligned}
p(\boldsymbol\beta_{t \ell}| \textbf{X}_{t \ell}, R_{t}, \sigma^{2}, T_{t}) & \propto p(R_{t}| \textbf{X}_{t \ell}, \boldsymbol\beta_{t \ell}, \sigma^{2}, T_{t}) p(\boldsymbol\beta_{t \ell}), \nonumber
\end{aligned}
$$
\noindent
which is a
$$
\begin{aligned}
\label{full_conditional_of_beta}
\textbf{N}_{q} \left(\boldsymbol\mu_{t \ell},  \sigma^{2} \boldsymbol\Lambda_{t \ell} \right),
\end{aligned}
$$
\noindent
where $\boldsymbol\mu_{t \ell} = \boldsymbol\Lambda_{t \ell} (\textbf{X}_{t \ell}^\top \textbf{r}_{t \ell})$, $\boldsymbol\Lambda_{t \ell} = (\textbf{X}_{t \ell}^\top \textbf{X}_{t \ell} + \textbf{V}^{-1})^{-1}$ and $\textbf{X}_{t \ell}$ is an $n_{t \ell} \times q$ matrix with all elements of the design matrix such that $i \in \mathcal{P}_{t \ell}$. The full conditional of $\sigma^{2}$ is similar to the expression in \eqref{BART_full_conditional_sigma2}, but with $\hat{y}_{i} = \sum_{t = 1}^{m} g(\textbf{x}_{i}; T_{t}, B_{t})$. Finally, the full conditional for $T_{t}$ is given by
$$
\begin{aligned}
p(T_{t}|\textbf{X}, R_{t}, \sigma^{2}) & \propto p(T_{t}) \int p(R_{t}| \textbf{X}, B_{t}, \sigma^{2}, T_{t}) p(B_{t})  d B_{t}, \\
& \propto p(T_{t}) p(R_{t}| \textbf{X}, \sigma^{2}, T_{t}),
\end{aligned}
$$
where
$$
\begin{aligned}
\label{full_conditional_of_T}
p(R_{t}& |\textbf{X},  \sigma^{2}, T_{t}) =  (\sigma^{2})^{-n/2} \mbox{ } \prod_{\ell=1}^{b_{t}} \left[  |\textbf{V}|^{-1/2} |\boldsymbol\Lambda_{t \ell}|^{1/2} \right. \nonumber \times \\
& \times \left. \exp \left( -\frac{1}{2 \sigma^{2}} \left[ - \boldsymbol\mu^{\top}_{t \ell} \boldsymbol\Lambda^{-1}_{t \ell} \boldsymbol\mu_{t \ell} + \textbf{r}_{t \ell}^\top \textbf{r}_{t \ell} \right] \right) \right]. \nonumber
\end{aligned}
$$

\begin{figure}
\centering
\begin{forest}
for tree={s sep=6mm}
[Root \\ $x_{2} < 10$
[$x_{1} < 0$ , edge label={node[midway,left,font=\scriptsize]{False}} [$\textbf{X}_{t1}\hat{\boldsymbol\beta}_{t1}$, edge label={node[midway,left,font=\scriptsize]{False}}] [$\textbf{X}_{t2}\hat{\boldsymbol\beta}_{t2}$ , edge label={node[midway,right,font=\scriptsize]{True}}]] [$x_{2} < 5$, edge label={node[midway,right,font=\scriptsize]{True}}
      [$\textbf{X}_{t3}\hat{\boldsymbol\beta}_{t3}$, edge label={node[midway,left,font=\scriptsize]{False}}]
      [$x_{3} < 5$, edge label={node[midway,right,font=\scriptsize]{True}},
[$\textbf{X}_{t4}\hat{\boldsymbol\beta}_{t4}$, edge label={node[midway,left,font=\scriptsize]{False}}]
         [$\textbf{X}_{t5}\hat{\boldsymbol\beta}_{t5}$ , edge label={node[midway,right,font=\scriptsize]{True}}]
      ]
] ]
\end{forest}
\caption{An example of a tree generated based on MOTR-BART. The quantities $x_{1}, x_{2}$ and $x_{3}$ represent covariates; $\textbf{X}_{t \ell}$ is a subset of the design matrix $\textbf{X}$ such that $i \in \mathcal{P}_{t \ell}$ and; $\hat{\boldsymbol\beta}_{t \ell} = (\hat{\beta}_{0t\ell}, \hat{\beta}_{1t\ell}, ...., \hat{\beta}_{p_{t \ell}t\ell})^{\top}$ is the parameter vector associated to the node $\ell$ of the tree $t$.}
\label{LM_BART_tree}
\end{figure}
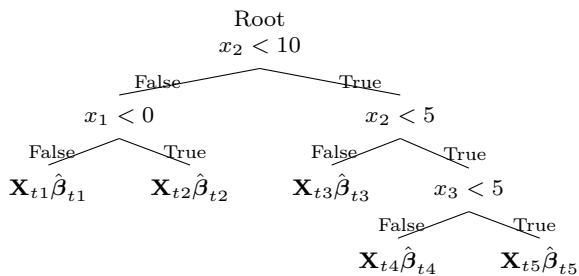

\noindent 
The main difference between BART and MOTR-BART can be seen in Figure \ref{LM_BART_tree}. Now, rather than having a constant as the predicted value for each terminal node, the prediction will be obtained from a linear predictor at node level. The purpose of introducing a linear predictor is to try to capture local linearity, reduce the number of trees and then possibly improve the prediction at node level.

The key point in MOTR-BART is which covariates should be considered in the linear predictor of each terminal node. At first glance, one might think that it would be advantageous to use variable selection techniques for regression models, such as ridge regression, Lasso \citep{tibshirani1996regression} or Horseshoe \citep{carvalho2010horseshoe}. Under the Bayesian perspective, these methods assume different priors on the regression coefficient vector and then estimate its components. In the ridge and Horseshoe regressions, a Gaussian with mean zero is assumed as the prior on the parameter vector. For Lasso regression, a Laplace distribution is considered. For MOTR-BART, we assume a Normal distribution with mean zero on $\boldsymbol\beta_{t \ell}$, but as the trees might change their dimensions depending on the moves growing and pruning, it is not possible to obtain the posterior distribution associated to each component of $\boldsymbol\beta_{t \ell}$ and then perform the variable selection.

Our idea to circumvent this issue is to consider in the linear predictor only covariates that have been used as a split in the corresponding tree. For instance, in Figure \ref{LM_BART_tree} three covariates are used as a split ($x_{1}$, $x_{2}$ and $x_{3}$). The plan is to include these three covariates in each of the five linear predictors. The intuition in doing so is that if a covariate has been utilised as a split, it means that it improves the prediction either because it has a linear or a non-linear relation with the response variable. If this relation is linear, this will be captured by the linear predictor. However, if the relation is non-linear, the coefficient associated to this covariate will be close to zero and the covariate will not have impact on the prediction.

We have also explored using only the ancestors of the terminal nodes in the linear predictor as well as replacing the uniform branching process, where the covariates are selected with equal probability, by the Dirichlet branching process proposed by \cite{linero2018bayesian}. To illustrate the first approach, we recall Figure \ref{LM_BART_tree}, where there are five terminal nodes and three covariates are used in the splitting rules. For the two left-most terminal nodes, only the covariates $x_{1}$ and $x_{2}$ would be considered in both linear predictors. For the terminal node 3, only $x_{2}$. For the right-most terminal nodes, $x_{3}$ and $x_{2}$ would be used. In relation to Linero's approach, rather than selecting the covariates with probability $1/p$, a Dirichlet prior is placed on the vector of splitting probability so that the covariates that are frequently used to create the internal nodes are more likely to be chosen. In the supplementary material, we show the performance of these and other strategies that we investigated to select the covariates that should be considered in the linear predictor.


\subsection{MOTR-BART for classification}
\label{LM_BART_classification}

The version of MOTR-BART that was presented in Section \ref{LM_BART} assumes that the response variable is continuous. In this Section, we provide the extension to the case when it is binary following the idea of \cite{chipman2010bart}, which used the strategy of data augmentation \citep{albert1993bayesian}. Firstly, we consider that $y_{i} \in \{ 0, 1\}$ and we introduce a latent variable 
$$
\begin{aligned}
z_{i} \sim \mbox{N}\left(\sum_{t=1}^{m} g(\textbf{x}_{i}, T_{t}, B_{t}), 1\right), \mbox{ with } i=1,..., n
\end{aligned}
$$

\noindent
such that $y_{i} = 1$ if $z_{i} > 0$ and $y_{i} = 0$ if $z_{i} \leq 0$. With this formulation, we have that $p(y_{i} = 1| \textbf{x}_{i}) = \mathrm{\Phi}(\sum_{t=1}^{m} g( \textbf{x}_{i}, T_{t}, B_{t}))$, where $\mathrm{\Phi}(\cdot)$ is the cumulative distribution function (cdf) of the standard Normal, which works as the link function that limits the output to the interval $(0,1)$. Here, there is no need to estimate the variance component as it is equal to 1. The priors on $T_{t}$ and $B_{t}$ are the same as in \eqref{Prior_T_t} and \eqref{Prior_beta_tl}, respectively. Finally, as the latent variable $z_{i}$ is introduced, it is necessary to compute its full conditional, which is given by
$$
\begin{aligned}
z_{i}| [y_{i} = 0] & \sim \mbox{N}_{(-\infty, 0)}\left(\sum_{t=1}^{m} g(\textbf{x}_{i}, T_{t}, B_{t}), 1\right), \\ 
z_{i}| [y_{i} = 1] & \sim \mbox{N}_{(0, \infty)}\left(\sum_{t=1}^{m} g(\textbf{x}_{i}, T_{t}, B_{t}), 1\right), \nonumber
\end{aligned}
$$
where $\mbox{N}_{(a,b)}(\cdot)$ denotes a truncated Normal distribution constrained to the interval $(a, b)$. Going back to Algorithm 1, some steps need to be modified or included:

\begin{enumerate}

    \item The update of $\sigma^{2}$ is no longer needed, because we set $\sigma^{2} = 1$; 
    
    \item The predicted values now consider the cdf of the standard Normal as a probit model in the form of $\hat{\textbf{y}}^{(k)} = \mathrm{\Phi} \left(\sum_{t=1}^{m} g(\textbf{X}, T_{t}, B_{t})\right)$;
    
    \item A Gibbs sampling step needs to be created to update the latent variables at each MCMC iteration. The update is done by drawing samples from $p(z_{i}| y_{i})$;
    
    \item Rather than calculating the partial residuals taking into account the response variable, we have that $R_{t}^{(k)} = \textbf{z}^{(k)} - \sum_{j \neq t}^{m} g(\textbf{X}, T_{j}, B_{j})$, where $\textbf{z}^{(k)}$ is the vector with the all latent variables at iteration $k$. For the first iteration, the vector $\textbf{z}^{(1)}$ needs to be initialised and $R_{1}^{(1)} = \textbf{z}^{(1)}$.
    
\end{enumerate}


\section{Results}
\label{Results}

In this Section, we compare MOTR-BART to BART, RF, GB, Lasso regression, soft BART, and LLF via simulation scenarios and real data applications using the Root Mean Squared Error (RMSE) as the accuracy measure. All results were generated by using R \citep{R} version 3.6.3 and the packages \texttt{dbarts} \citep{chipman2010bart}, \texttt{ranger} \citep{ranger}, \texttt{gbm} \citep{gbm}, \texttt{glmnet} \citep{friedman2010regularization}, \texttt{SoftBart} \citep{SoftBart}, and \texttt{grf} \citep{grf}. We use the default behaviour of these packages, except where otherwise specified below. We also tried running the Linear Random Forests (LRF, \cite{kunzel2019linear}) algorithm. However, we got errors when using the \texttt{forestry} R package and then we decided not to consider the LRF in our comparisons.

Throughout this Section, we present results for two versions of our method. The first one is MOTR-BART (10 trees), which uses the Dirichlet branching process and estimates $\tau_{\beta_{0}}$ and $\tau_{\beta}$, while the second is MOTR-BART (10 trees, fixed var), which uses the uniform branching process and sets $\tau_{b} = m$. As a default version, we recommend the MOTR-BART (10 trees).


\subsection{Simulation}
\label{Simulation}
To compare the algorithms, we simulate data from the equation proposed by \cite{friedman1991multivariate}. This data set is widely used in testing tree-based models and has been used repeatedly to evaluate the performance of BART and extensions \citep{friedman1991multivariate, chipman2010bart, linero2018bayesian}. We generate the response variable considering five covariates via:
$$
\begin{aligned}
y_{i} = 10 \mbox{sin}(\pi x_{i1} x_{i2}) + 20 (x_{i3} - 0.5)^{2} + 10 x_{i4} + 5 x_{i5} + \epsilon_{i},
\end{aligned}
$$
where the covariates $x_{ip} \sim \mbox{U}(0,1)$, with $p = 1, \ldots, 5$, and $\epsilon_{i} \sim \mbox{N}(0,1)$. For the simulation, we created 9 data sets with different numbers of observations (200, 500 and 1000) and covariates (5, 10 and 50). For those scenarios with 10 and 50 covariates, the additional $x$ values do not have any impact on the response variable.

Each simulated data set was split into 10 different training ($80\%$) and test ($20\%$) partitions. For MOTR-BART, 10 trees were considered, 1000 iterations as burn-in, 5000 as post-burn-in, $\texttt{alpha} = \alpha = 0.95$ and $\texttt{beta} = \beta = 2$. To choose the number of trees (10) for MOTR-BART, we initially tested a range of possible values, such as 3, 10 and 50, and then we used cross-validation to select that setting that presented the lowest RMSE. The set up for \texttt{dbarts} was similar to MOTR-BART, except for the number of trees (10 and 200, the default). For the packages $\texttt{ranger}$ and $\texttt{gbm}$, the default options were kept, except for the number of trees (200) and the parameter \texttt{interaction.depth = 3}. For the $\texttt{glmnet}$, we followed the manual and used a 10-fold cross-validation with $\texttt{type.measure = `mse'}$ to obtain the estimate of the regularisation parameter $\texttt{lambda.min}$, which is the value that minimises the cross-validated error under the loss function chosen in $\texttt{type.measure}$. As in \cite{chipman2010bart}, we evaluate the convergence of MOTR-BART and BART by eye from the plot of $\sigma^{2}$ after the burn-in period.

\begin{figure}
\includegraphics[height=0.9\textwidth]{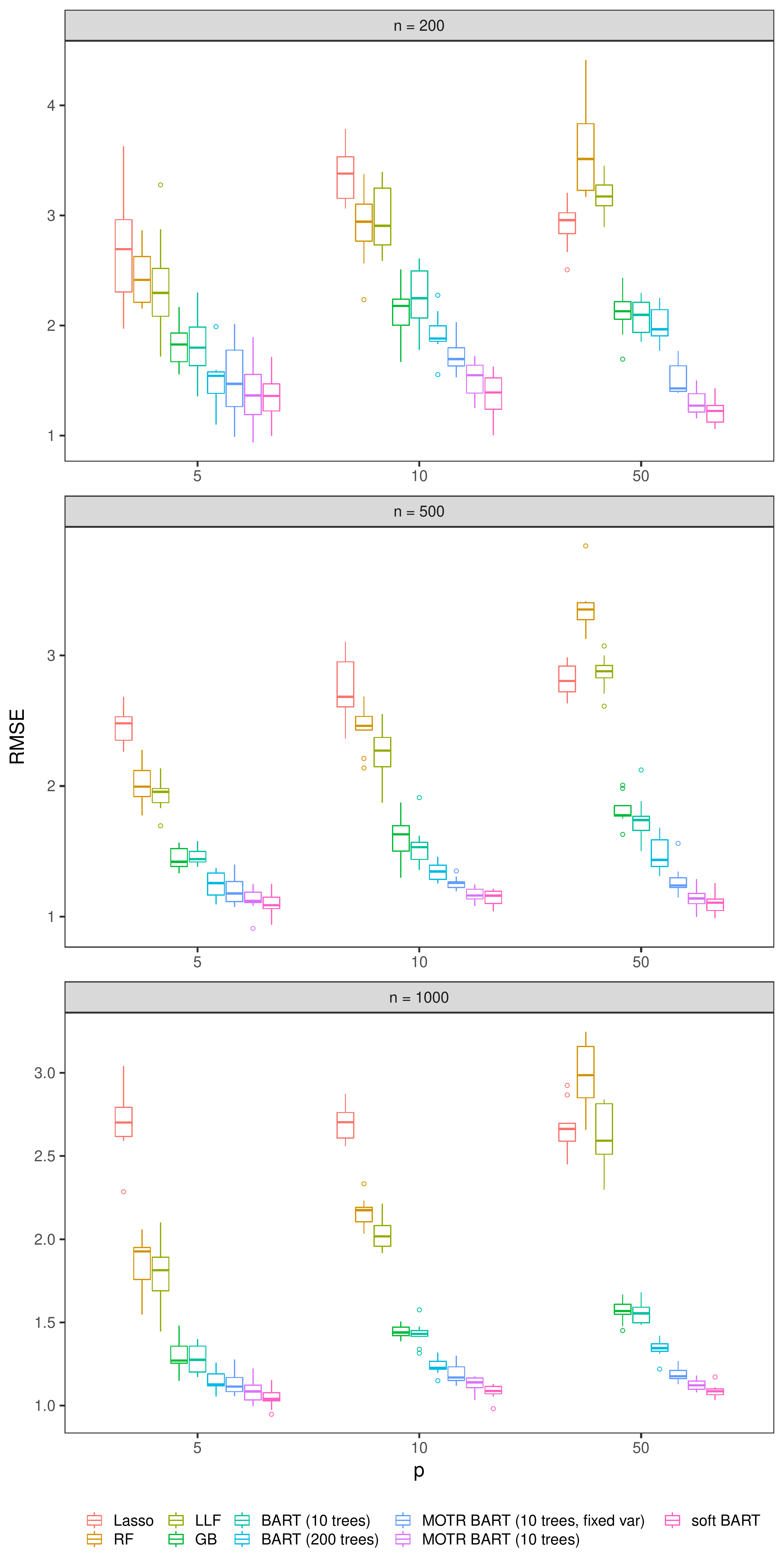}
\caption{Comparison of RMSE for Friedman data sets on test data for different combinations of $n$ (200, 500 and 1000) and $p$ (5, 10 and 50).}
\label{Simulation_test}       
\end{figure}

In Figure \ref{Simulation_test}, we present the comparison of the algorithms MOTR-BART, BART, RF, GB, Lasso, soft BART, and LLF in terms of RMSE on test data. Note that we have BART (10 trees) and BART (200 trees; default). The first version considers 10 trees and was run to see how BART would perform with the equivalent number of trees of MOTR-BART. We can see that for different combinations of number of observations ($n$) and covariates ($p$), soft BART and MOTR-BART (10 trees) consistently presented the best results for all scenarios. When compared to both versions of the original BART, both versions of MOTR-BART present lower median values of RMSE and slightly greater variability. However, the variability reduces as $n$ and $p$ increase. Further, we notice that MOTR-BART (10 trees) benefits from penalising the intercepts and slopes differently. In addition, it is possible to observe that the number of noisy covariates impacts on the performance of RF and LLF. For all values of $n$, their RMSEs increase with the number of covariates.

To further analyse the improvements given by MOTR-BART over standard BART, in Appendix A we present Table \ref{Friedman_terminal_mean}, which shows the mean of the total number of terminal nodes utilised for BART to calculate the final prediction taking into account all the 5000 iterations. The idea of Table \ref{Friedman_terminal_mean} is to show that MOTR-BART has similar or better performance whilst using fewer parameters than standard BART. As the default version of BART defaults to 200 trees, which is far more trees than MOTR-BART uses, we created Table \ref{Friedman_terminal_mean} to highlight that although MOTR-BART estimates fewer parameters, it still remains competitive to the default BART. For MOTR-BART, we consider the mean of the number of parameters estimated in the linear predictors. As BART and soft BART predict a constant for each terminal node, the number of `parameters' estimated is equal to the number of terminal nodes. On the other hand, MOTR-BART estimates an intercept, which is equivalent to the constant that BART predicts, plus the parameters associated to those covariates that have been used as a split in the corresponding tree. For instance, if a tree has 5 terminal nodes and 2 numeric variables are used in the splitting rules, MOTR-BART will estimate 15 parameters.  For BART, we set the argument \texttt{keepTrees = TRUE} and then we extracted from the sampler object \texttt{fit} the content of \texttt{getTrees()}. For both MOTR-BART and BART, we firstly summed the number of parameters for all trees along the MCMC iterations and then averaged it over the 10 sets.

In Table \ref{Friedman_terminal_mean}, we can observe, for example, for the Friedman data set with $n = 1000$ and $p = 50$ that BART (10 trees) utilised 212,421 parameters on average to calculate the final prediction, while MOTR-BART (10 trees), BART (200 trees), and soft BART used 391,193, 2,371,140, and 255,155, respectively. For all simulated data sets, MOTR-BART presented lower RMSE than BART (200 trees), even though it estimates far fewer parameters. From Table \ref{Friedman_terminal_mean}, it is possible to obtain the mean number of terminal nodes per tree by dividing the column `Mean' by the number of MCMC iterations (5000) times the number of trees (10 or 200). In this case, we note that both versions of BART produce small trees with the mean number of terminal nodes per tree varying between 2 and 5. Due to the greater number of trees, BART (200 trees) has the lowest mean, regardless the number of observations and covariates. In contrast, MOTR-BART has the mean number of parameters per tree varying from 5 to 8. Comparatively speaking, this is somewhat expected once MOTR-BART estimates a linear predictor. In this way, the trees from MOTR-BART tend to be shallower than those from BART (10 trees), but with more parameters estimated overall. It is important to highlight that the numbers from BART and MOTR-BART cannot be compared to those from RF, as the former work with the residuals and the latter with the response variable itself. The numbers for GB are not shown as the quantity of terminal nodes in each tree is fixed due to the parameter settings \texttt{interaction.depth = 3}.

In our simulations, MOTR-BART utilised just 10 trees and its results were better than RF, GB, BART (10 and 200 trees) and LLF. In practice, different number of trees may be compared via cross-validation and hence a choice can be made such that the cross-validated error is minimised.


\subsection{Application}
\label{Application}
In this Section, we compare the predictive performance of MOTR-BART to RF, GB, BART, soft BART, and LLF in terms of RMSE on four real data sets. The first one (Ankara) has 1,609 rows and contains weather information for the city of Ankara from 1994 to 1998. The goal is to predict the mean temperature based on 9 covariates. The second is the Boston Housing data set, where the response variable is the median value of properties in suburbs of Boston according to the 1970 U.S. census. This data set has 506 rows and 18 explanatory variables. The third data set (Ozone) has 330 observations and 8 covariates and is about ozone concentration in Los Angeles in 1976. The aim is to predict the amount of ozone in parts per million (ppm) based on wind speed, air temperature, pressure gradient, humidity and other covariates. The fourth data set (Compactiv) refers to a multi-user computer that had the time of its activity measured under different tasks. The goal is to predict the portion of time that the computer runs in user mode for 8,192 observations based on 21 covariates. These data sets are a subset of 9 sets considered by \cite{bartMachine}.

As with the Friedman data, we consider two versions of BART (10 and 200 trees) and MOTR-BART (10 trees and 10 trees, fixed var), and we split the data into 10 different train ($80\%$) and test ($20\%$) sets. Furthermore, no transformations were applied to the response variables and all results are based on the test data.

\begin{figure}
\includegraphics[height=0.9\textwidth]{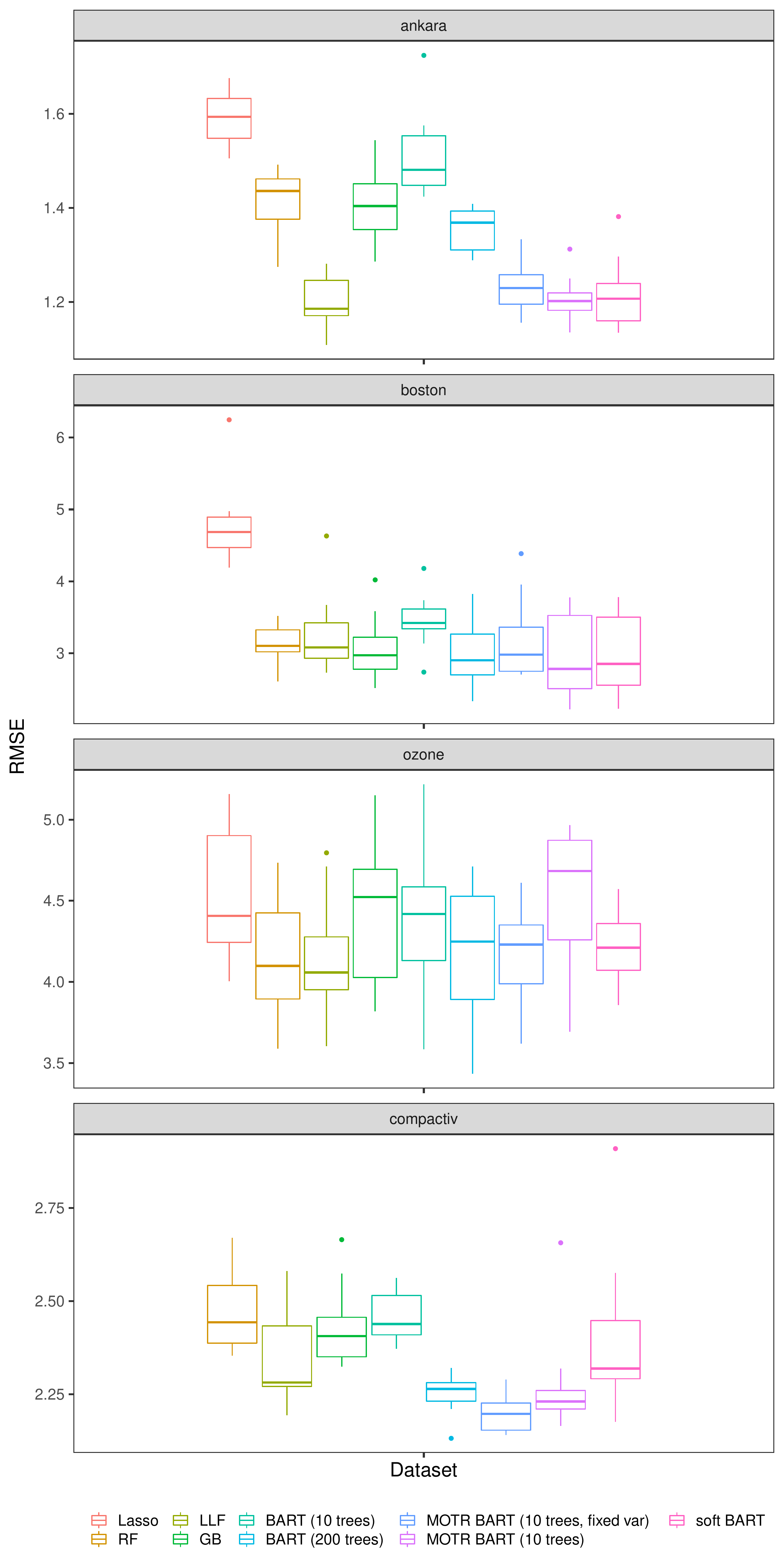}
\caption{Comparison of RMSE for Ankara, Boston, Ozone and Compactiv data sets on test data.}
\label{real_data}       
\end{figure}

Figure \ref{real_data} shows the results of RMSE on test sets. It is possible to note that MOTR-BART (10 trees) presents the lowest or second lowest median RMSE on all data sets, except for Ozone. For Ankara, RF and GB have quite similar results and Lasso presents the highest RMSE. For Boston, Lasso regression shows the highest RMSE, while MOTR-BART (10 trees) and soft BART do not differ much in terms of median and quartiles. For Ozone, it can be seen that MOTR-BART (10 trees) presents the highest RMSE and that LLF, RF and MOTR-BART (10 trees) have the lowest median values. For Compactiv, RF and GB show similar results, while MOTR-BART (10 trees, fixed var) presents the lowest RMSE. To facilitate the visualisation, the results for Lasso are not shown for the data set Compactiv, as it has RMSEs greater than 9. In Appendix B, however, Table \ref{tab_real_data_RMSE_rank} reports the median and the first and third quartiles of the RMSE for all algorithms and data sets.

In Table \ref{Real_data_mean_terminal_nodes} (see Appendix B), we show the mean of the total number of parameters/terminal nodes created for BART to generate the final prediction for each data set. For MOTR-BART, the numbers correspond to the mean of the total of parameters estimated. For instance, for the data set Ankara, 304,696 terminal nodes were used on average by BART (10 trees), while BART (200 trees) estimated 2,250,599 and MOTR-BART 546,959. As can be seen, MOTR-BART estimates more parameters than BART (10 trees) for all data sets, as we expect. However, when compared to BART (200 trees), MOTR-BART utilises significantly fewer parameters to obtain similar or better performance, except for Compactiv.


\section{Discussion}

In this paper, we have proposed an extension of BART, called MOTR-BART, that can be seen as a combination of BART and Model Trees. In MOTR-BART, rather than having a constant as predicted value for each terminal node, a linear predictor is estimated considering only those covariates that have been used as a split in the corresponding tree. Furthermore, MOTR-BART is able to capture linear associations between the response and covariates at node level as well as it requires fewer trees to achieve equivalent or better performance when compared to other methods.

Via simulation studies and real data applications, we showed that MOTR-BART is highly competitive when compared to BART, Random Forests, Gradient Boosting, Lasso Regression, soft BART and LLF. In simulation scenarios, MOTR-BART outperformed the other tree-based methods, except soft BART. In the real data applications, four data sets were considered and MOTR-BART provided great predictive performance.

Due to the structure of MOTR-BART, to evaluate variable importance or even to select the covariates that should be included in the linear predictors is not straightforward. Recall that Model Trees was introduced in the context of one tree, where statistical methods of variable selection, such as Forward, Backwards or Stepwise can be performed at node level. Compared to other tree-based methods that consider only one tree, Model Trees produces much smaller trees \citep{landwehr2005logistic}, which helps to alleviate the computational time required by the variable selection procedures. In theory, one might think that it would be possible to use such a procedure for MOTR-BART, but in practice they would be a burden as they would be performed for each terminal node out of all trees within every MCMC iteration.

In the Bayesian context, \cite{chipman2010bart} propose to use the inclusion probability as a measure of variable importance. Basically, this metric is the proportion of times that a covariate is used as a split out of all splitting rules over all trees and MCMC iterations. However, this measure gives us an idea about the covariates that are important for the splitting rules and does not say anything about which covariates that should be included in the linear predictors.

In this sense, the variable selection/importance remains as a challenge that may be investigate in future work, once conventional procedures are not suitable. One might try proposing adaptations of ridge, lasso or horseshoe regressions for trees. Another extension could be replacing the linear functions by Splines to provide even further flexibility and capture local non-linear behaviour, which is a subject of ongoing work. Finally, Model Trees can be incorporated to other BART extensions, such as BART for log-linear models \citep{jared2017}, soft BART \citep{lineroAnDyang2018}, and BART for log-normal and gamma hurdle models \citep{linero2018semiparametric}. We hope to produce an R package that implements our methods shortly; current code is available at \url{https://github.com/ebprado/MOTR-BART}.


\begin{acknowledgements}
We thank the editors and the two anonymous referees for their comments that greatly improved the earlier version of the paper. Estev\~ao Prado’s work was supported by a Science Foundation Ireland Career Development Award grant number 17/CDA/4695 and SFI research centre 12/RC/2289P2. Andrew Parnell’s work was supported by: a Science Foundation Ireland Career Development Award (17/CDA/4695); an investigator award (16/IA/4520); a Marine Research Programme funded by the Irish Government, co-financed by the European Regional Development Fund (Grant-Aid Agreement No. PBA/CC/18/01); European Union’s Horizon 2020 research and innovation programme under grant agreement No 818144; SFI Centre for Research Training 18CRT/6049, and SFI Research Centre awards 16/RC/3872 and 12/RC/2289P2.
\end{acknowledgements}


\section*{Appendix A: Simulation results}
\label{Appendix_A}

In this Section, we present results related to the simulation scenarios shown in Subsection \ref{Simulation}. In total, 9 data sets were created based on Friedman's equation considering some combinations of sample size ($n$) and number of covariates ($p$). In Tables \ref{tab:1} and \ref{tab:12}, the medians and quartiles of the RMSE are shown for the algorithms MOTR-BART, BART, GB, RF, Lasso, soft BART, and LLF. The values in this table were graphically shown in Figure \ref{Simulation_test}. In addition, Table \ref{Friedman_terminal_mean} presents the mean number of parameters utilised by BART, MOTR-BART, and soft BART to calculate the final prediction.

\begin{table}
\caption{The median of the RMSE on test data of the Friedman data sets when $n = 200 \mbox{ and } 500$. The values in parentheses are the first and third quartiles, respectively.}
\label{tab:1}       
\centering
\begin{tabular}{lll}
  \hline
Algorithm & $p$ & RMSE \\ 
  \hline
  \multicolumn{ 3}{c}{\textbf{$n = 200$}} \\ \hline
  MOTR-BART & 5 & \textbf{1.36 (1.19;1.55)} \\
  MOTR-BART (fixed var) & 5 & 1.47 (1.26;1.78) \\ 
  BART (10 trees) & 5 & 1.80 (1.63;1.99) \\ 
  BART (200 trees) & 5 & 1.54 (1.38;1.58) \\ 
  GB & 5 & 1.83 (1.67;1.93) \\ 
  RF & 5 & 2.41 (2.21;2.63) \\
  Lasso & 5 & 2.69 (2.30;2.96) \\ 
  soft BART & 5 &\textbf{1.36 (1.22;1.47)} \\
  LLF & 5 & 2.30 (2.08;2.52) \\
  \hline
  MOTR-BART & 10 & \textbf{1.55 (1.39;1.64)} \\ 
 MOTR-BART (fixed var) & 10 & 1.70 (1.63;1.80) \\ 
  BART (10 trees) & 10 & 2.25 (2.07;2.49) \\ 
  BART (200 trees) & 10 & 1.88 (1.86;2.00) \\ 
  GB & 10 & 2.18 (2.00;2.24) \\ 
  RF & 10 & 2.94 (2.76;3.10) \\ 
  Lasso & 10 & 3.38 (3.15;3.53) \\ 
  soft BART & 10 & \textbf{1.39 (1.24;1.52)} \\ 
  LLF & 10 & 2.91 (2.73;3.25) \\ 
  \hline
  MOTR-BART & 50 & \textbf{1.27 (1.21;1.38)} \\ 
  MOTR-BART (fixed var) & 50 & 1.43 (1.40;1.63) \\ 
  BART (10 trees) & 50 & 2.10 (1.94;2.21) \\ 
  BART (200 trees) & 50 & 1.97 (1.90;2.14) \\ 
  GB & 50 & 2.13 (2.06;2.22) \\ 
  RF & 50 & 3.51 (3.23;3.83) \\ 
  Lasso & 50 & 2.96 (2.84;3.02) \\ 
  soft BART & 50 & \textbf{1.22 (1.12;1.27)} \\ 
  LLF & 50 & 3.17 (3.09;3.28) \\ 
    \hline
  \multicolumn{ 3}{c}{\textbf{$n = 500$}} \\ \hline
  MOTR-BART & 5 & \textbf{1.12 (1.11;1.19)} \\ 
  MOTR-BART (fixed var) & 5 & 1.18 (1.11;1.27) \\ 
  BART (10 trees) & 5 & 1.44 (1.42;1.50) \\ 
  BART (200 trees) & 5 & 1.26 (1.17;1.33) \\ 
  GB & 5 & 1.42 (1.38;1.52) \\ 
  RF & 5 & 2.00 (1.92;2.12) \\ 
  Lasso & 5 & 2.48 (2.35;2.53) \\ 
  soft BART & 5 & \textbf{1.09 (1.06;1.15)} \\ 
  LLF & 5 & 1.96 (1.87;1.98) \\ 
  \hline
  MOTR-BART & 10 & \textbf{1.16 (1.14;1.21)} \\ 
  MOTR-BART (fixed var) & 10 & 1.26 (1.22;1.27) \\ 
  BART (10 trees) & 10 & 1.53 (1.44;1.57) \\ 
  BART (200 trees) & 10 & 1.35 (1.28;1.39) \\ 
  GB & 10 & 1.63 (1.50;1.70) \\ 
  RF & 10 & 2.46 (2.43;2.53) \\ 
  Lasso & 10 & 2.68 (2.61;2.95) \\ 
  soft BART & 10 & \textbf{1.16 (1.10;1.19)} \\ 
  LLF & 10 & 2.27 (2.15;2.37) \\
  \hline
  MOTR-BART & 50 & \textbf{1.14 (1.10;1.18)} \\ 
  MOTR-BART (fixed var) & 50 & 1.24 (1.22;1.30) \\ 
  BART (10 trees) & 50 & 1.74 (1.66;1.77) \\ 
  BART (200 trees) & 50 & 1.43 (1.38;1.59) \\ 
  GB & 50 & 1.78 (1.77;1.85) \\ 
  RF & 50 & 3.35 (3.27;3.40) \\ 
  Lasso & 50 & 2.80 (2.72;2.92) \\ 
  soft BART & 50 & \textbf{1.11 (1.05;1.13)} \\ 
  LLF & 50 & 2.88 (2.83;2.92) \\
   \hline
\end{tabular}
\end{table}

\begin{table}
\caption{The median of the RMSE on test data of the Friedman data sets when $n = 1000$. The values in parentheses are the first and third quartiles, respectively.}
\label{tab:12}       
\centering
\begin{tabular}{lll}
  \hline
Algorithm & $p$ & RMSE \\ 
  \hline
  \multicolumn{ 3}{c}{\textbf{$n = 1000$}} \\ \hline
  MOTR-BART & 5 & \textbf{1.09 (1.03;1.12)} \\ 
  MOTR-BART (fixed var) & 5 & 1.11 (1.08;1.17) \\ 
  BART (10 trees) & 5 & 1.28 (1.20;1.36) \\ 
  BART (200 trees) & 5 & 1.13 (1.12;1.19) \\ 
  GB & 5 & 1.27 (1.25;1.36) \\ 
  RF & 5 & 1.93 (1.76;1.95) \\ 
  Lasso & 5 & 2.70 (2.62;2.79) \\ 
  soft BART & 5 & \textbf{1.04 (1.03;1.08)} \\ 
  LLF & 5 & 1.81 (1.69;1.89) \\
  \hline
  MOTR-BART & 10 & \textbf{1.14 (1.11;1.17)} \\ 
  MOTR-BART (fixed var) & 10 & 1.17 (1.15;1.23) \\ 
  BART (10 trees) & 10 & 1.43 (1.42;1.45) \\ 
  BART (200 trees) & 10 & 1.23 (1.22;1.27) \\ 
  GB & 10 & 1.44 (1.42;1.47) \\ 
  RF & 10 & 2.17 (2.10;2.19) \\ 
  Lasso & 10 & 2.70 (2.61;2.76) \\ 
  soft BART & 10 & \textbf{1.09 (1.07;1.12)} \\ 
  LLF & 10 & 2.02 (1.96;2.08) \\
  \hline
  
  MOTR-BART & 50 & \textbf{1.12 (1.10;1.15)} \\ 
  MOTR-BART (fixed var) & 50 & 1.18 (1.16;1.21) \\ 
  BART (10 trees) & 50 & 1.55 (1.50;1.59) \\ 
  BART (200 trees) & 50 & 1.35 (1.33;1.37) \\ 
  GB & 50 & 1.57 (1.55;1.61) \\ 
  RF & 50 & 2.99 (2.85;3.16) \\ 
  Lasso & 50 & 2.66 (2.59;2.70) \\
  soft BART & 10 & \textbf{1.09 (1.07;1.10)} \\ 
  LLF & 10 & 2.59 (2.51;2.81) \\
   \hline
\end{tabular}
\end{table}

\begin{table}[ht]
\caption{Friedman data sets: Mean and standard deviation of the total number of terminal nodes created for BART and soft BART to generate the final prediction over 5,000 iterations. For MOTR-BARTs, the values correspond to the mean of the total number of parameters estimated in the linear predictors.}
\label{Friedman_terminal_mean}
\centering
\begin{tabular}{llrr}
  \hline
Algorithm & $p$ & Mean & Std \\ 
  \hline
  \multicolumn{ 4}{c}{\textbf{$n = 200$}} \\ \hline
  MOTR-BART & 5 & 302,447 & 18,210 \\ 
  MOTR-BART (fixed var) & 5 & 263,079 & 11,990 \\ 
  BART (10 trees) & 5 & 163,468 & 7,393 \\ 
  BART (200 trees) & 5 & 2,468,707 & 6,734 \\ 
  soft BART & 5 & 250,615 & 6,599 \\ 
  \hline
  MOTR-BART  & 10 & 326,678 & 20,309 \\ 
  MOTR-BART (fixed var) & 10 & 258,380 & 13,530 \\ 
  BART (10 trees) & 10 & 145,458 & 7,380 \\ 
  BART (200 trees) & 10 & 2,470,333 & 3,670 \\ 
  soft BART & 10 & 256,391 & 5,577 \\ 
  \hline
  MOTR-BART & 50 & 327,751 & 28,627 \\ 
  MOTR-BART (fixed var) & 50 & 251,469 & 12,376 \\ 
  BART (10 trees) & 50 & 134,809 & 4,447 \\ 
  BART (200 trees) & 50 & 2,428,259 & 5,368 \\ 
  soft BART & 50 & 256,184 & 6,754 \\ 
  \hline
  \multicolumn{ 4}{c}{\textbf{$n = 500$}} \\ \hline
  MOTR-BART & 5 & 364,786 & 21,978 \\ 
  MOTR-BART (fixed var) & 5 & 364,258 & 17,478 \\ 
  BART (10 trees) & 5 & 203,625 & 7,950 \\ 
  BART (200 trees) & 5 & 2,470,900 & 8,739 \\ 
  soft BART & 5 & 257,769 & 6,727 \\ 
  \hline
  MOTR-BART & 10 & 382,528 & 24,768 \\ 
  MOTR-BART (fixed var) & 10 & 354,755 & 25,238 \\ 
  BART (10 trees) & 10 & 206,394 & 8,694 \\ 
  BART (200 trees) & 10 & 2,448,212 & 9,171 \\ 
  soft BART & 10 & 256,465 & 2,829 \\ 
  \hline
  MOTR-BART & 50 & 384,828 & 18,099 \\ 
  MOTR-BART (fixed var) & 50 & 330,434 & 33,566 \\ 
  BART (10 trees) & 50 & 178,779 & 8,700 \\ 
  BART (200 trees) & 50 & 2,407,661 & 14,314 \\ 
  soft BART & 50 & 254,164 & 9,334 \\ 
  \hline
  \multicolumn{ 4}{c}{\textbf{$n = 1000$}} \\ \hline
  MOTR-BART & 5 & 389,479 & 23,247 \\ 
  MOTR-BART (fixed var) & 5 & 396,280 & 29,040 \\ 
  BART (10 trees) & 5 & 271,878 & 8,977 \\ 
  BART (200 trees) & 5 & 2,425,863 & 8,276 \\ 
  soft BART & 5 & 257,656 & 9,881 \\ 
  \hline
  MOTR-BART  & 10 & 410,517 & 30,346 \\ 
  MOTR-BART (fixed var) & 10 & 390,274 & 22,442 \\ 
  BART (10 trees) & 10 & 256,511 & 7,812 \\ 
  BART (200 trees) & 10 & 2,415,372 & 8,575 \\ 
  soft BART & 10 & 255,604 & 6,549 \\ 
  \hline
  MOTR-BART & 50 & 391,193 & 16,127 \\ 
  MOTR-BART (fixed var) & 50 & 380,365 & 40,069 \\ 
  BART (10 trees) & 50 & 212,421 & 5,959 \\ 
  BART (200 trees) & 50 & 2,371,140 & 14,287 \\ 
  soft BART & 50 & 255,155 & 4,400 \\ 
   \hline
\end{tabular}
\end{table}


\newpage
\section*{Appendix B: Real data results}
\label{Appendix_B}

This Appendix presents two tables with results associated to the data sets Ankara, Boston, Ozone and Compactiv. In Table \ref{tab_real_data_RMSE_rank}, it is reported the median and quartiles of the RMSE computed on 10 test sets. The values in this table are related to the Figure \ref{real_data} from Subsection \ref{Application}. Further, Table \ref{Real_data_mean_terminal_nodes} shows the mean number of parameters utilised by BART, MOTR-BART, and soft BART to calculate the final prediction for the aforementioned data sets.

\setlength{\tabcolsep}{4pt} 
\renewcommand{\arraystretch}{1} 
\begin{table}[ht]
\caption{Real data sets: Comparison of the median RMSE (and first and third quartiles) for Ankara, Boston, Ozone, and Compactiv data sets on test data. The acronym `fv' stands for `fixed var'.}
\label{tab_real_data_RMSE_rank}
\centering
\begin{tabular}{llll}
  \hline
Data set & Algorithm & RMSE & rank \\ 
  \hline
\multirow{8}{*}{Ankara} & MOTR-BART & \textbf{1.20 (1.18;1.22)} & 2\\ 
 & MOTR-BART (fv) & 1.23 (1.20;1.26) & 4\\ 
 & BART (200 trees) & 1.37 (1.31;1.39) & 5 \\ 
 & BART (10 trees) & 1.48 (1.45;1.55) & 8 \\ 
 & GB & 1.40 (1.35;1.45) & 6 \\ 
 & RF & 1.44 (1.38;1.46) & 7 \\ 
 & Lasso & 1.59 (1.55;1.63) & 9\\ 
 & soft BART &1.21 (1.16;1.24) & 3 \\ 
 & LLF & \textbf{1.19 (1.17;1.25)} & 1\\ 

  \hline
  \multirow{8}{*}{Boston} & MOTR-BART & \textbf{2.78 (2.51;3.53)} & 1\\ 
   & MOTR-BART (fv) & 2.98 (2.75;3.36) & 5\\ 
   & BART (200 trees) & 2.90 (2.70;3.27) & 3 \\ 
   & BART (10 trees) & 3.42 (3.34;3.62) & 8 \\ 
   & GB & 2.97 (2.78;3.22) & 4 \\ 
   & RF & 3.10 (3.02;3.33) & 6\\ 
   & Lasso & 4.69 (4.47;4.89) & 9 \\ 
   & soft BART & \textbf{2.85 (2.56;3.50)} & 2 \\ 
   & LLF & 3.08 (2.93;3.42) & 7 \\ 
  \hline
  \multirow{8}{*}{Ozone} & MOTR-BART & 4.68 (4.26;4.87) & 9\\ 
   & MOTR-BART (fv) & 4.23 (3.99;4.35) & 3\\ 
   & BART (200 trees) & 4.25 (3.89;4.53) & 4\\ 
   & BART (10 trees) & 4.42 (4.13;4.59) & 6 \\ 
   & GB & 4.52 (4.03;4.69) & 8 \\ 
   & RF & \textbf{4.10 (3.89;4.43)} & 2 \\ 
   & Lasso & 4.41 (4.24;4.90) & 7\\ 
   & soft BART & 4.21 (4.07;4.36) & 5 \\ 
   & LLF & \textbf{4.06 (3.95;4.28)} & 1\\ 
  \hline
  \multirow{8}{*}{Compactiv} 
   & MOTR-BART & \textbf{2.23 (2.21;2.26)} & 2 \\ 
   & MOTR-BART (fv) & \textbf{2.20 (2.15;2.23)} & 1 \\ 
   & BART (200 trees) & 2.26 (2.23;2.28) & 3\\ 
   & BART (10 trees) & 2.44 (2.41;2.51) & 7 \\ 
   & GB & 2.41 (2.35;2.46) & 6 \\ 
   & RF & 2.44 (2.39;2.54) & 8 \\ 
   & Lasso & 9.97 (9.51;10.09) & 9\\ 
   & soft BART & 2.32 (2.29;2.45) & 5\\ 
   & LLF & 2.28 (2.27;2.43) & 4\\ 
   \hline
\end{tabular}
\end{table}

\setlength{\tabcolsep}{4pt} 
\renewcommand{\arraystretch}{1} 
\begin{table}[ht]
\caption{Real data sets: Mean and standard deviation of the total number of terminal nodes created for BART and soft BART to generate the final prediction over 5,000 iterations. For MOTR-BARTs, the values correspond to the mean of the total number of parameters estimated in the linear predictors. The acronym `fv' stands for `fixed var'.}
\label{Real_data_mean_terminal_nodes}
\centering
\begin{tabular}{llrr}
  \hline
Data set & Algorithm & Mean & Std \\ 

  \hline
  \multirow{4}{*}{Ankara} & MOTR-BART & 546,959 & 36,977 \\ 
  & MOTR-BART (fv) & 485,743 & 40,840 \\ 
   & BART (10 trees) & 304,696 & 8,872 \\ 
   & BART (200 trees) & 2,250,599 & 11,798 \\ 
   & soft BART & 312,927 & 11,539 \\ 
  \hline
  \multirow{4}{*}{Boston} & MOTR-BART & 748,468 & 58,945 \\ 
  & MOTR-BART (fv) & 414,762 & 50,705 \\ 
   & BART (10 trees) & 204,038 & 10,143 \\ 
   & BART (200 trees) & 2,389,130 & 14,244 \\ 
  & soft BART & 318,171 & 17,078 \\ 
  \hline
  \multirow{4}{*}{Ozone} & MOTR-BART & 272,370 & 42,809 \\ 
  & MOTR-BART (fv) & 182,189 & 8,093 \\
   & BART (10 trees) & 137,239 & 2,642 \\ 
   & BART (200 trees) & 2,343,350 & 5,128 \\ 
   & soft BART & 268,948 & 5,667 \\ 
  \hline
  \multirow{4}{*}{Compactiv} & MOTR-BART & 2,990,494 & 298,221 \\ 
  & MOTR-BART (fv) & 1,529,666 & 102,940 \\ 
   & BART (10 trees) & 539,621 & 15,759 \\ 
   & BART (200 trees) & 2,649,167 & 29,989 \\ 
   & soft BART & 711,860 & 49,087 \\ 
  \hline
\end{tabular}
\end{table}


\newpage
\bibliographystyle{spbasic}       
\bibliography{template}   

\end{document}